\documentclass[sigconf]{acmart}
\usepackage{times}
\usepackage{url}
\usepackage{latexsym}
\usepackage{amsmath}
\usepackage{booktabs} 
\usepackage{graphicx}
\usepackage{multibib}
\usepackage{bbm}
\usepackage[labelformat=empty]{subfig}
\expandafter\def\expandafter\UrlBreaks\expandafter{\UrlBreaks
  \do\a\do\b\do\c\do\d\do\e\do\f\do\g\do\h\do\i\do\j%
  \do\k\do\l\do\m\do\n\do\o\do\p\do\q\do\r\do\s\do\t%
  \do\u\do\v\do\w\do\x\do\y\do\z\do\A\do\B\do\C\do\D%
  \do\E\do\F\do\G\do\H\do\I\do\J\do\K\do\L\do\M\do\N%
  \do\O\do\P\do\Q\do\R\do\S\do\T\do\U\do\V\do\W\do\X%
  \do\Y\do\Z}
 
\usepackage{soul}
\makeatletter
\def\@copyrightspace{\relax}
\makeatother
\usepackage{etoolbox}
\makeatletter
\patchcmd{\maketitle}{\@copyrightspace}{}{}{}
\makeatother
\renewcommand\footnotetextcopyrightpermission[1]{} 
\setcopyright{none}
\acmConference{}{}{}

\title{Modeling ``Newsworthiness'' for Lead-Generation Across Corpora}

\author{Alexander Spangher
     \qquad Nanyun Peng 
     \qquad Jonathan May
     \qquad Emilio Ferrara
     \\
   Information Sciences Institute / University of Southern California \\
   {\tt \{spangher, peng, jonmay, ferrarae\}@isi.edu} \\
}

\date{}

\begin{document}
\begin{abstract}
    Journalists obtain ``leads'', or story ideas, by reading large corpora of government records: court cases, proposed bills, etc. However, only a small percentage of such records are interesting documents. We propose a model of ``newsworthiness'' aimed at surfacing interesting documents. We train models on automatically labeled corpora -- published newspaper articles -- to predict whether each article was a front-page article (i.e., \textbf{newsworthy}) or not (i.e., \textbf{less newsworthy}). We transfer these models to unlabeled corpora -- court cases, bills, city-council meeting minutes -- to rank documents in these corpora on ``newsworthiness''. A fine-tuned RoBERTa model achieves .93 AUC  performance on heldout labeled documents, and .88 AUC on expert-validated unlabeled corpora. We provide interpretation and visualization for our models.
\end{abstract}

\maketitle

\section{Introduction}

The overwhelming amount of digital information available in the internet age, ``information overload'' \cite{white2000confronting}, has been well-known to cause decision fatigue \cite{goswami2015analysing}, anxiety \cite{bawden2009dark} and declines in learning \cite{mayer2001cognitive}. Although solutions to information overload have been well explored in the literature \cite{koltay2017bright,ho2001towards,hagel1999private}, current approaches require either user-adaptation, algorithmic refinements, or introducing ``infomediaries'' like a portal or community. These approaches either assume that the user knows the kind of information they are seeking, require human intervention, or need an observable metric to optimize (like engagement). None of these approaches, further, promote the kind of general information consumption that leads to shared facts \cite{lee2018social}.


The primary task of journalism, on the other hand, is the identification and publication of interesting, \textit{general} pieces of information, i.e., information that is \textit{newsworthy}. It is, in principle, the task of ``creating an informed electorate'': ``newsworthiness'', classically, refers to information that informs voters in a democracy \cite{mcintyre2016makes}. However, judging the newsworthiness of a piece of information requires intensive human efforts, based on intuition about what kind of information is important for voters, and so far has been challenging to replicate algorithmically (Section \ref{scn:relwork}).

To this end, we offer a narrower, operational definition of ``newsworthiness'' that seeks to interpret and apply historical expert judgements. Our definition is: \textit{how likely is this piece of information to appear on the \ul{front page} of a major newspaper?} With this definition, we propose a simple classification task: is this record substantially similar to news articles that have appeared on the front page?

\begin{table}
    \centering
    \begin{tabular}{|p{1.5cm}|p{6cm}|}
    	\toprule
    	\textbf{Corpus} & \textbf{Top predictions} \\
    	\toprule
    	City Council Meeting Min. & \textit{Rules} which prohibit use of funds for 2026 World Cup unless U.S. Soccer Fed. provides equitable pay to U.S. Women's and U.S. Men's Team \\
    	\midrule
    	State Bills & \textit{Bill} requiring school districts to participate in Medicaid for health and social services. \\
    	\midrule
    	Court Filings & \textit{Indictment} returned against Governor Rick Perry for ... exercising authority to veto appropriations vested in the Governor by Texas Constitution. \\
    	\bottomrule
    \end{tabular}
    \caption{{\small Sample of top newsworthy records, ranked by RoBERTa \cite{liu2019roberta}.}}
\end{table}

In this work, we train models to learn ``newsworthiness'' by classifying the page that newspaper articles are published on. We use these models to sort documents in other corpora used by journalists. We formalize a method for ranking content based on an observable metric that captures decades of journalistic and editorial judgement. Our core  contributions  are: (1) a formulation of the ``newsworthiness'' problem setup, (2) the introduction of expert-annotated corpora, novel to researchers but commonly used by journalists (totaling millions of documents with hundreds of expert annotations, which we will release) (3) a demonstration that ``newsworthiness'' can be learned from one corpora and transferred more broadly.

Formalizing a ranking of content based on newsworthiness is essential to helping us identify democratically relevant information. 
We offer in this work a proof-of-concept of a simple principle for filtering and sorting information. We see potential applications for such predictions in online algorithmic ranking systems: while web layouts on major social media and content sites are presently defined algorithmically to maximize short-term signals (i.e. click-through rates) or long-term signals (i.e. subscriber potential), a ``newsworthiness'' ranking could construct more socially relevant rankings. More fundamentally, we are pose question: can the archives of human decisions made, over the decades, by experts practiced in the art of sorting necessary information in print help us inform new approaches for doing so on the web?

We also see such an approach helping journalists filter information, alerting journalists when a particularly newsworthy document is published in any of the corpora used 
-- e.g., court cases, city council minutes -- 
in their day-to-day work to find story ideas, check powerful individuals and 
keep abreast of the workings of government. Declining newspaper revenues and deep staff cuts have left many journalist institutions without the ability to thoroughly evaluate many records, especially on a state and local-level \cite{miller2018news}. 
Such an advancement could allow journalists to reduce time spent finding stories, sift through more corpora, find the most interesting documents. 

Our research goals are two-fold:
\begin{itemize}
	\item We train models on a labeled corpora -- the \textit{New York Times} annotated corpus, 1987 - 2007 -- to predict whether articles are \textbf{front-page} or \textbf{not}.
	\item We transfer these models to unlabeled corpora -- city council minutes, law text, court cases -- to rank documents by ``newsworthiness''.
\end{itemize}

\section{Problem Description}
\label{scn:prob-desc}

Our goal is to model $p(\text{newsworthy} | \text{text})$ for any input text. We consider two sets of corpora: a set of labeled corpora $C^{'} = \{c^{'}_i\}$ and a set of unlabeled $C^{''} = \{c^{''}_k\}$. For each document $a \in C^{'}$, we have labels for whether the article was published on the front page or not, for document $d \in C^{''}$, we do not. Our predictive tasks are: we seek to build models that (1) accurately classify labels on $C_.^{'}$, which we evaluate on a held-out set. And (2) generalizes to $C_.^{''}$, which we will evaluate using expert annotation.

\section{Data}
\label{scn:data}
\begin{figure}
	\includegraphics[width=\linewidth]{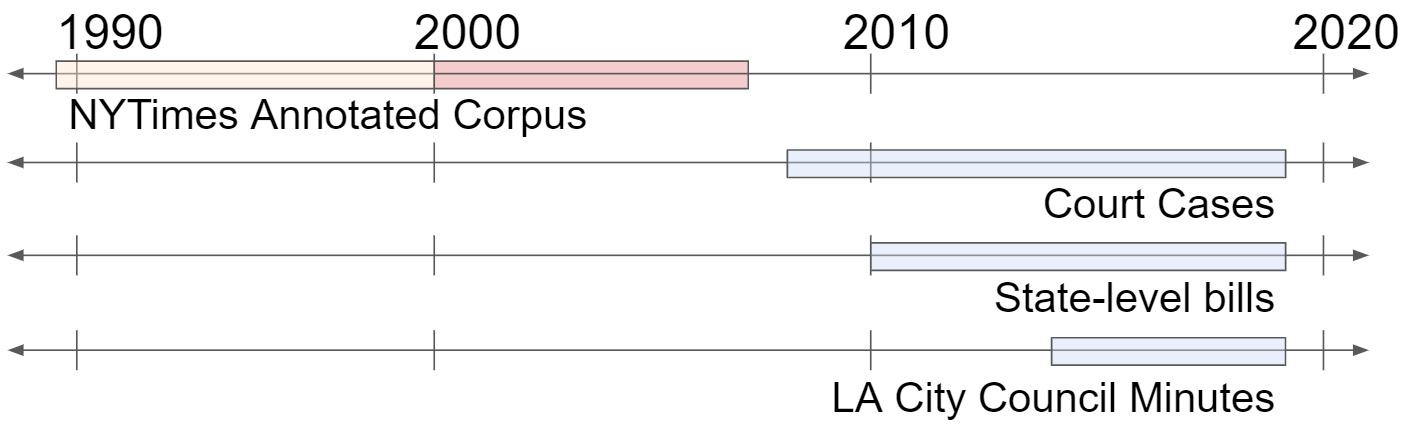}
    \caption{Timelines of the corpora gathered. The \textit{New York Times} Annotated Corpus is our labeled corpus. The red block denotes the test split. The rest of the corpora is chosen to be non-overlapping in time.}
	\label{fig:data-daterange}
\end{figure}

We collect four corpora (one labeled and three unlabeled), shown in Figure \ref{fig:data-daterange}.

\noindent\textbf{$C_1^{'}$: New York Times Annotated Corpus (1987-2007)}
We use the \textit{New York Times} Annotated Corpus\footnote{\url{catalog.ldc.upenn.edu/LDC2008T19}} as training and evaluation, which contains 1.8 million articles published from 1980--2007, ($40,000$ front-page, or $A1$). Each article has a number of attributes including: headline, full-text and page-number. Length of articles are between 300-1200 words, with a median of 800 words. We do not consider articles published on Saturday and Sunday, as front-page weekend articles tend to be longer, more narrative and generally less ``hard news'' than weekday articles\footnote{Based on personal interview with newspaper layout editors.}.

\begin{figure}
    \centering
    \includegraphics[width=.8\linewidth]{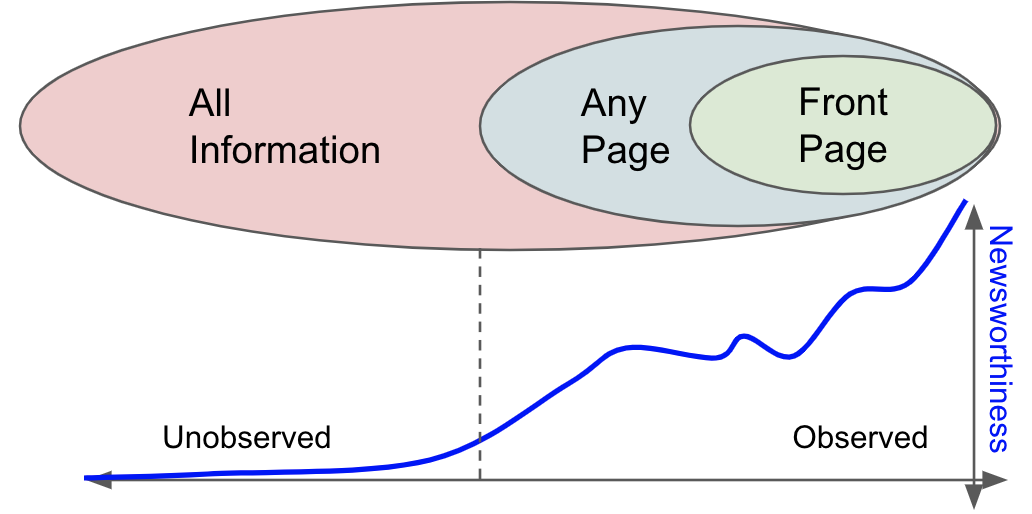}
    \caption{Limitations of our work: We do not observe learn a function characterizing newsworthiness of information not included in the newspaper at all, nor do we acknowledge that different non-front pages of the newspaper may be more newsworthy than others.}
    \label{fig:limitations}
\end{figure}

\noindent\textbf{$C_2^{''}$: Los Angeles City Council Meeting Minutes (2015-2019)}

The Los Angeles City Council publishes summaries of topics discussed in council meetings online\footnote{\url{cityclerk.lacity.org/lacityclerkconnect/}}. We scrape all city council meetings occurring between 2015-2019. For each meeting, we parse the separate agenda items, which consist of a title, a case-number, and a brief description of the item. There are between 10-20 agenda items per meeting. In total, we collect $113,000$ documents. 

City council meetings were once a venue for journalists seeking to provide coverage to their local communities. Newsworthy city council meetings can address local issues of concern in cities across  the  nation, like  homelessness, can also involve a local solution to a  national problem, like pay equity.

\noindent\textbf{$C_3^{''}$: State-level Bills (2010-2018)} We collect all state-level bills passed between 2010-2018, as recorded by Open Secrets\footnote{\url{openstates.org/}}. In total, we collect $1.04$ million documents. The information provided includes title, a brief description, and subject-tags. 

Like city-council meetings, state law was once broadly covered by journalists. However, while coverage gaps are not as egregious as on the city-council level, there still might be instances where state-law is newsworthy, but is missed by journalism outlets.

\noindent\textbf{$C_4^{''}$: Opinions from Appeals Court Cases. (2008-2018)} We use CourtListener\footnote{\url{courtlistener.com/}}, an open-source scraper that compiles court-dockets from different state and federal court houses, to collect all documents produced by appeals cases, totaling over $110,000$ documents. For each document, we have the full-text of the opinions (including defense and judgement.) Court records have long been used by journalists in their daily reporting, as extensive information about the parties involved becomes part of the public record
\footnote{A journalist once described to me his work investigating Exxon Mobile:
\begin{quote}
    I wasn't making any leeway until I saw that some environmental advocacy lawyers brought Exxon to court. They used the deposition process to put secret company documents into the public record. From these documents, I learned a lot about the inner workins of the company, their efforts to suppress climate change science, and their plans.
\end{quote}
}

We choose non-overlapping date-ranges for $C^{'}$ and $C^{''}$ (shown in Figure \ref{fig:data-daterange}). We do this to minimize the change that articles in $C^{'}$ directly reference documents in $C^{''}$, as we want to rank-order $C^{''}$ on general newsworthy patterns rather than specific topics/people. Our corpora vocabularies exhibit different levels of divergence (Figure \ref{fig:kl-divergence}). The highest divergence is observed between $C^{'}_1$ and $C^{''}_2$: \textbf{City Council} is the most local of our corpora. The train/test splits show a low KL-divergence of $.1$ in both directions.

\section{Experimental Design and Results}
We preprocess all of our corpora to eliminate a list of stopwords specific to newspaper publishing\footnote{Ex) ``op-ed'', ``sportsmonday'', ``business review''. We identify this list through iteratively training LR and examining top coefficients.} 
We train four text classifiers: Logistic Regression with BOW thresholds (min\_df=.01, max\_df=.5, vocab\_size=13,000) (LR) \cite{scikit-learn}, FastText\cite{joulin2016bag}, pretrained Bert-Base (BT),\cite{devlin2018bert}, and pre-trained RoBERTa (RT) \cite{liu2019roberta}, on a balanced training set of $C_1^{'}$ articles published in 1987-2001 ($y_{1}^{train} = y_{0}^{train} \approx 45,000$ articles). \textit{\ul{We use AUC as a metric because we are most interested in the rank-order of documents that our classifiers generate.}}

\begin{table}
	\begin{tabular}{|l|r|r|}
		\toprule
		{} & Full-Text & Extracted Events \\
		\midrule
		\textbf{Log. Reg. (LR)} & .85 & .69\\
		\textbf{FastText (FT)} & .88 & .81 \\
		\textbf{BERT-base (BT)} & .91 & .71 \\
		\textbf{RoBERTa-base (RT)} & .93 & .76 \\
		\bottomrule
	\end{tabular}
	\caption{{\small Eval. 1: AUC on $C^{'}$. First row shows training/testing on full text of article, second on extracted events only.}}
	\label{tbl:task-1-res}
\end{table}


\noindent\textbf{Goal 1: Performance on $C^{'}$ heldout} To test if our models accurately predict newsworthiness, we evaluate on an unbalanced training set of $C^{'}$ articles published in 2001-2007 ($y_1^{test} \approx 11,000$ articles, $y_0^{test} \approx 210,000$ articles.) As shown in the top row of Table \ref{tbl:task-1-res}, the top-scoring model is RT ($.93$ AUC).

\begin{figure}
    \includegraphics[width=.7\linewidth]{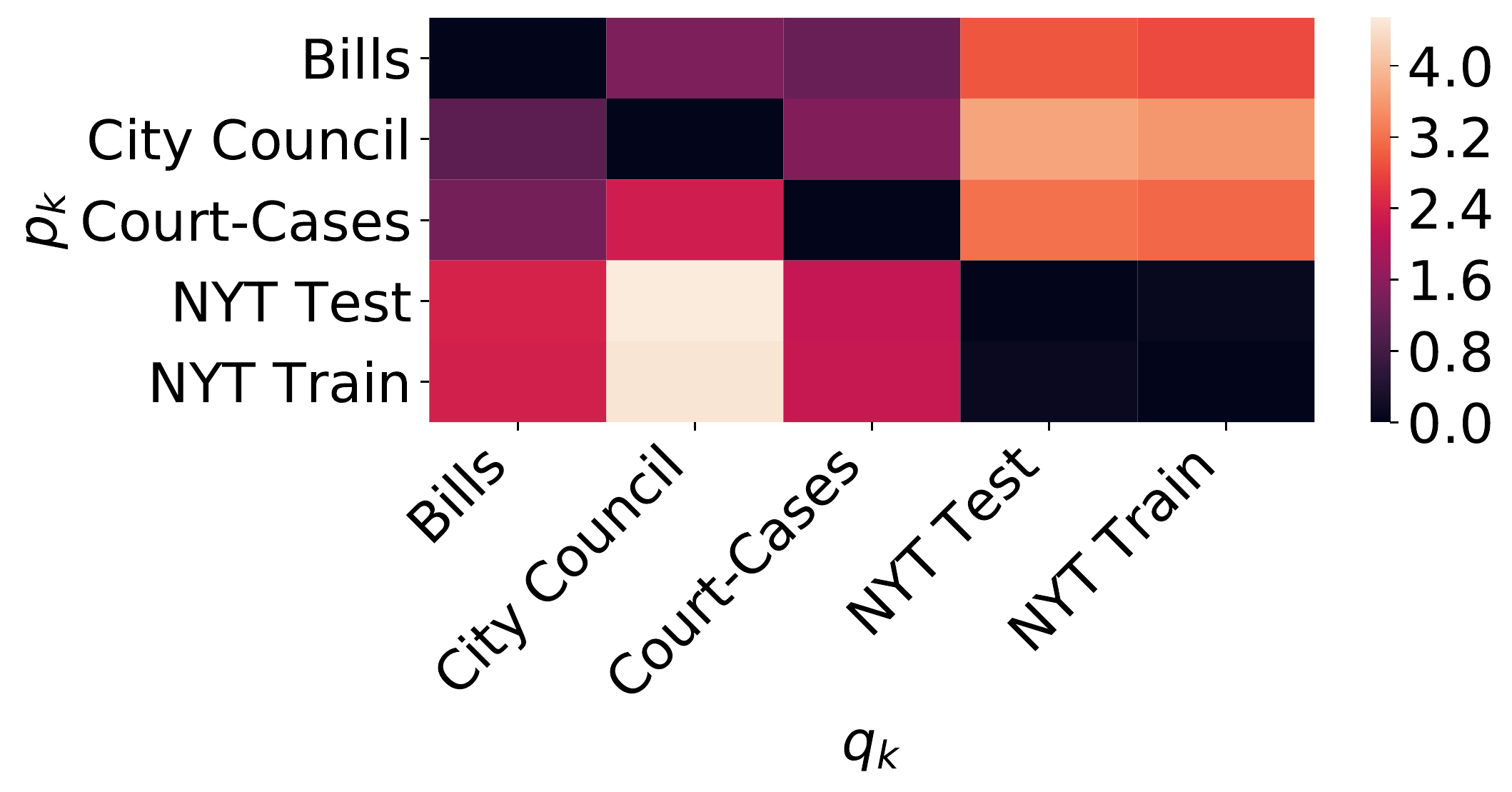}
    \caption{Variation between the corpora, shown with KL-Divergence over unigram counts for each corpus. The Y-axis shows the base distribution, $p_k$ and the X-axis shows the comparison distribution, $q_k$.
    }
    \label{fig:kl-divergence}
\end{figure}

\begin{figure}
	\centering
    \includegraphics[width=.6\linewidth]{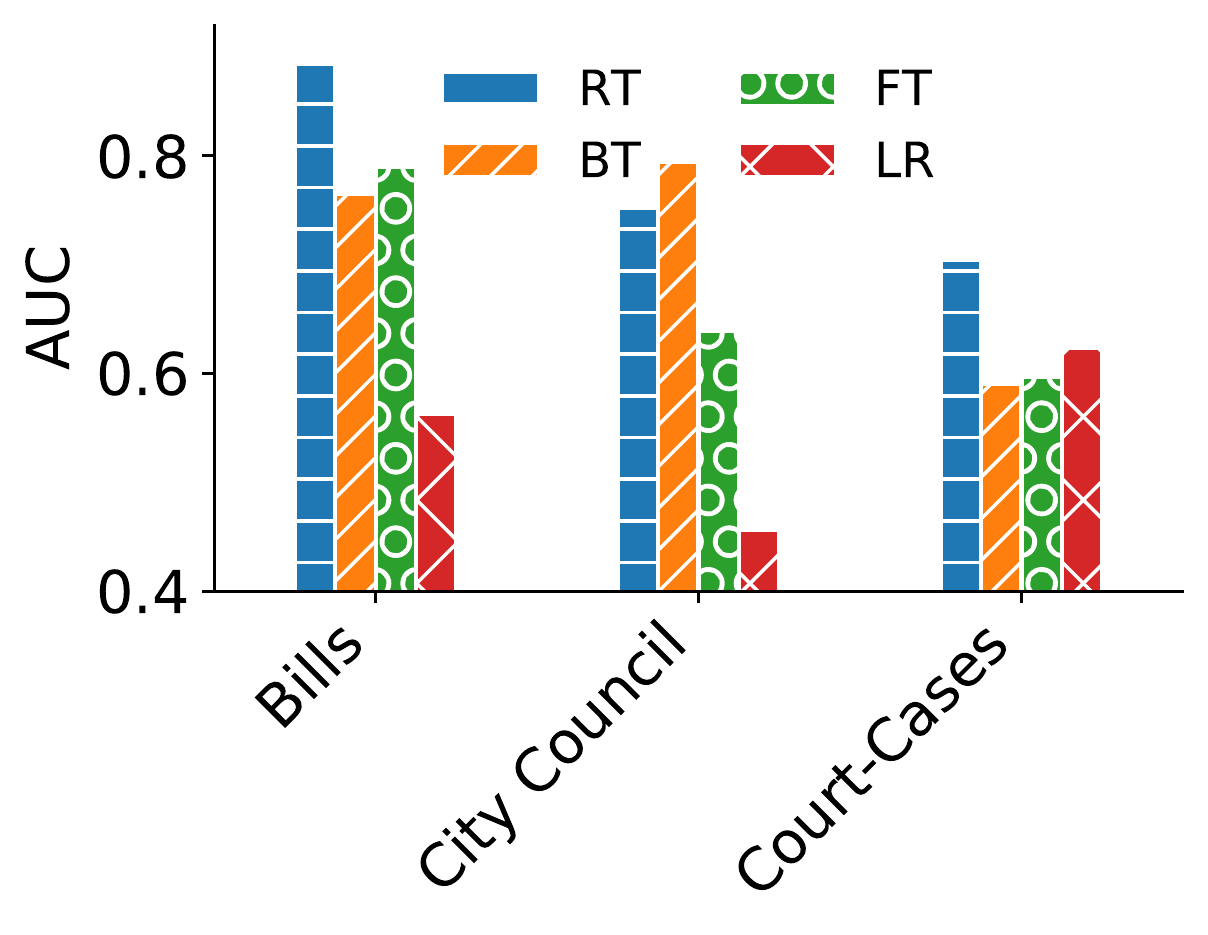}
    \caption{AUC of model-transfer on the external corpora gathered.}
    \label{fig:auc-annotation}
\end{figure}

\noindent\textbf{Goal 2: Performance on $C^{''}$} For our second task, an expert annotator\footnote{Our annotator worked at major national newspaper and was present for page-layout decisions.} rates $100$ documents from each unlabeled corpora in blind trials with a simple rating $\in \{0, 1\}$ for whether they would assign a journalist to investigate a story based on the document. We report these results in Figure \ref{fig:auc-annotation}. The RT model still outperforms for the Bills corpora (.88 AUC) and the Court-Cases corpora (.70 AUC), which according to Figure \ref{fig:kl-divergence} are more similar to the training data. The BT model is the top performer for the City Council corpora (.79 AUC). We explore possible explanations in Section \ref{scn:explanatory-results}. 


\section{Limitations}

Our problem formulation and choice of data are clearly a limited view into the realm of newsworthy information. Firstly, as shown in Figure \ref{fig:limitations}, we only train our models on information that made it into the newspaper in the first place. Thus, we only learn to distinguish ``newsworthiness'' on a limited set of information and assume that this transfers into unseen parts of the function. Secondly, we neglect that different non-front page pages might have different newsworthy value, like section fronts. Thirdly, we do not account for interdependencies between articles: i.e. ``heavy news days'' vs. ``light news days''. We seek to mitigate some these problems by relying on a professional journalist to annotate our transfer accuracy, but we also demonstrate, via the accuracy of our models, that such limitations need not halt preliminary research into this problem.

\section{Explanatory Results}
\label{scn:explanatory-results}
\begin{figure*}[t]
    \centering
	\includegraphics[width=.26\linewidth, height=.85\linewidth, angle=90]{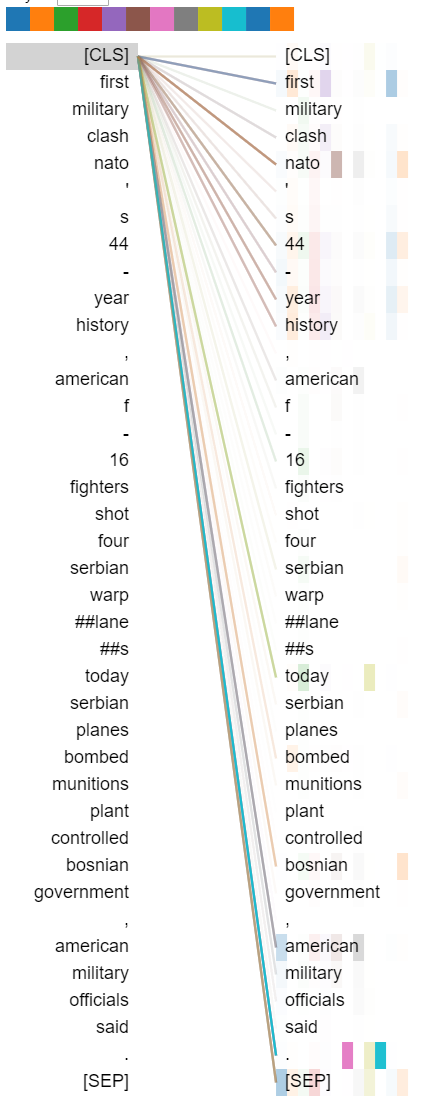}
	\caption{{\small Attention weights given by BERT model for example sentence from \textit{New York Times} corpus appearing on the front page. Attention to $[CLS]$ label shown. Each color represents $1$ of $12$ attention heads.}}
	\label{fig:attn-diagram}
\end{figure*}

We first hypothesized that the models were learning to discriminate newsworthy events. To test this, we performed event extraction as in \cite{han2019joint} and concatenated event arguments and anchors and ran our models just on this text. However, as shown in the second row of Table \ref{tbl:task-1-res}, models run on extracted events underperformed model on the full-text of articles (or first 512 word-pieces for BT, RT).

Clearly, other aspects of the text were contributing signal. We show as an illustration the attention scores given by BT to the $[CLS]$ token, in Figure \ref{fig:attn-diagram}. The model attends to the immediacy word, ``today'', which represent late-breaking forms of newsworthy events, and the historical word, ``first''. To further explore the attention given by BT, we show in Table \ref{tbl:bert-attn} the top average attention scores, $A_{BT}$, across heads in the last layer for all documents in $C{'}$-test. $A_{BT}$ where $y=1$ have \textbf{emotional salience}, like ``threatens'' \cite{ostgaard1965factors}. In contrast, $A_{BT}$ where $y=0$ are mainly \textbf{place-based}. Local news is often published in the Metro Section: BT learns to distinguish emotional salience and locality. 
We run a similar experiment on RT, shown in Table \ref{tbl:rt-attn}. $A_{RT}$ where $y=1$ are \textbf{time-based} words. This is another form of newsworthiness: when events are time-stamped, they are more likely to be important in the short-term \cite{galtung1965structure}. $A_{RT}$ where $y=0$ are international signifiers. The dynamics are parallel to BT: international news is often published in the International section. Finally, we show coefficients for LR, which are all more \textbf{topic-based}. Top positive $\beta^{+}$ are political events, while top $\beta^{-}$ are business terms.\footnote{The phrase ``survived wife'' is often used in Obituaries.}
Overall, it appears our models are each learning different aspects of newsworthiness. We do no hypothesize BT or RT were well-suited for each aspect, just that there are many aspects of newsworthiness that each model independently converged on. However, we leave a more careful analysis of different kinds of newsworthiness to future work.

\begin{table}
	\centering
	\begin{tabular}{|p{2cm}r|p{1.7cm}r|}
		\toprule
		\multicolumn{2}{|c}{\textbf{Front Page}} & \multicolumn{2}{c|}{\textbf{Not Front Page}}\\
		\toprule
		Word &   Atten. &          Word &   Atten. \\
		\midrule
		{\small threatens} &  {\small .059} &        {\small suffolk} & {\small  .078} \\
		{\small startling} &  {\small .053}  &          {\small diary} &  {\small .060} \\
		{\small follows} &  {\small .053} &    {\small connecticut} &  {\small .053}  \\
		{\small stunned} &  {\small .052} &         {\small knicks} &  {\small .049} \\
		\bottomrule
	\end{tabular}
	\caption{Words receiving top attention weights to the \textit{[CLS]} token in the final layer BT.}
	\label{tbl:bert-attn}
\end{table}

\begin{table}
	\centering 
	\begin{tabular}{|p{2cm}r|p{1.7cm}r|}
		\toprule
		\multicolumn{2}{|c}{\textbf{Front Page}} & \multicolumn{2}{c|}{\textbf{Not Front Page}}\\
		\toprule
		Word &   Atten. &          Word &   Atten. \\
		\midrule
		{\small December} &   {\small .019}  &            {\small Vatican}  &    {\small .019} \\
		{\small February} &   {\small .019} &            {\small editor}     &    {\small .017} \\
		{\small point} &        {\small .019} &          {\small Korean}   &    {\small .017} \\
		{\small governments}  & {\small .018} &           {\small Beijing}  &    {\small .017} \\
		\bottomrule
	\end{tabular}
	\caption{Top attention weights to the \textit{[CLS]} token in the final layer RT.}
	\label{tbl:rt-attn}
\end{table}

\section{Related Work}
\label{scn:relwork}

The challenge we explore in this work relates to a subfield of computational journalism called \textit{lead generation}, or identifying pieces information that could lead to news articles \cite{cohen2011computational}.\footnote{\textit{Computational journalism} is an emerging field aimed at identifying applications of statistical and computational approaches to impact the traditional journalistic practice.}

Of existing approaches to lead-generation, one is given by \cite{diakopoulos2010diamonds}, who seeks to quantify a piece of content's \textit{relevance} to a given topic, its \textit{uniqueness}, and its \textit{sentiment}. The authors develop metrics for such attributes and applies them to surfacing tweets related to presidential speeches. Their system is designed to surface tweets made during newsworthy speeches, like the president's State of the Union address, and filter tweets to those that are relevant and unique. It then facilitates exploration of these tweets through an interface. 
This approach is useful for generating leads for events journalists know to search for, like political speeches, but are limited in identifying new topics of coverage. Our approach does not place such a constraint on journalists and can surface novel content, independent of preconceived topic.
A second approach is \textit{anomaly detection}. Systems like Newsworthy analyze open-source, numerical datasets, like polling data and housing market data, to discover outliers \cite{newsworthy_slideshow}, which are surfaced to journalists to investigate. Such approaches interpret ``newsworthiness'' as a deviation from the mean, which imposes distributional assumptions on datasets observed (and also assumes that datapoints closer to the mean are \textit{not} newsworthy). 
This approach might be relevant for data-driven stories, but would not capture many event-driven stories. Our approach surfaces textual data, thus operates in a different domain.
%
A third approach involves fact-checking: systems like ClaimBuster scan news, speeches and social media for claims being made by politicians. Once claims are identified, they are forwarded to journalists to check \cite{adair2017progress}. While this approach applies to a subset of news concerned with fact-checking, it is not clear that this has wide-applicability to the most newsworthy stories.
All three of these approaches utilize specific, expert-designed metrics for newsworthy content. Our approach sidesteps these systems and directly models newsworthiness. 

\begin{table}

        \begin{tabular}{|p{2.2cm}r|p{2.2cm}r|}
    		\toprule
    		\multicolumn{2}{|c}{\textbf{Top Pos. Coef.}} & \multicolumn{2}{c|}{\textbf{Top Neg. Coef.}}\\
    		\toprule
    		Word &      $\beta$ &                          Word &      $\beta$ \\
    		\midrule
    		{\small nation largest} &        {\small .25} &          {\small share earns} &     {\small -.41} \\
    		{\small people killed} &         {\small .25} &          {\small survived wife} &   {\small -.41} \\
    		{\small communist party} &       {\small .23} &          {\small media business} &  {\small -.37} \\
    		{\small court ruled} &    {\small .23} &                 {\small share} &           {\small -.37} \\
    		\bottomrule
    	\end{tabular}
    	\caption{N-grams receiving the top positive and negative coefficients in the Logistic Regression model.}
    	\label{tbl:lr-coef}
\end{table}

\section{Conclusion}

In this work, we have formalized a novel classification task, ``newsworthiness ranking'', for which ample training data exists. We have translated a human judgement on the democratic importance of information into an observable metric and modeled it with high accuracy. We provide three novel datasets to demonstrate transfer potential. We have explored factors contributing to historical judgements of newsworthiness. Such exploration, we observe, has the potential to contribute positively to our information economy by helping both readers and journalists find and consume more socially relevant information.

\clearpage
\bibliography{acl2019}

\begin{thebibliography}{1}
\expandafter\ifx\csname natexlab\endcsname\relax\def\natexlab#1{#1}\fi

\bibitem[{Diakopoulos et~al.(2010)Diakopoulos, Naaman, and
  Kivran-Swaine}]{diakopoulos2010diamonds}
Nicholas Diakopoulos, Mor Naaman, and Funda Kivran-Swaine. 2010.
\newblock Diamonds in the rough: Social media visual analytics for journalistic
  inquiry.
\newblock In \emph{2010 IEEE Symposium on Visual Analytics Science and
  Technology}, pages 115--122. IEEE.

\end{thebibliography}


\begin{thebibliography}{10}

\bibitem{TeXFAQ}
{UK \TeX{} Users Group}.
\newblock {UK} list of {\TeX} frequently asked questions.
\newblock \url{https://texfaq.org}, 2019.

\bibitem{Downes04:amsart}
Michael Downes and Barbara Beeton.
\newblock {\em The \textsf{amsart}, \textsf{amsproc}, and \textsf{amsbook}
  document~classes}.
\newblock American Mathematical Society, August 2004.
\newblock \url{http://www.ctan.org/pkg/amslatex}.

\bibitem{Fiorio15}
Cristophe Fiorio.
\newblock {\em {a}lgorithm2e.sty---package for algorithms}, October 2015.
\newblock \url{http://www.ctan.org/pkg/algorithm2e}.

\bibitem{Brito09}
Rog\'erio Brito.
\newblock {\em The algorithms bundle}, August 2009.
\newblock \url{http://www.ctan.org/pkg/algorithms}.

\bibitem{Heinz15}
Carsten Heinz, Brooks Moses, and Jobst Hoffmann.
\newblock {\em The Listings Package}, June 2015.
\newblock \url{http://www.ctan.org/pkg/listings}.

\bibitem{Fear05}
Simon Fear.
\newblock {\em Publication quality tables in {\LaTeX}}, April 2005.
\newblock \url{http://www.ctan.org/pkg/booktabs}.

\bibitem{ACMIdentityStandards}
Association for Computing Machinery.
\newblock {\em {ACM} Visual Identity Standards}, 2007.
\newblock \url{http://identitystandards.acm.org}.

\bibitem{Sommerfeldt13:Subcaption}
Axel Sommerfeldt.
\newblock {\em The subcaption package}, April 2013.
\newblock \url{http://www.ctan.org/pkg/subcaption}.

\bibitem{Nomencl}
Boris Veytsman, Bern Schandl, Lee Netherton, and C.~V. Radhakrishnan.
\newblock {\em A package to create a nomenclature}, September 2005.
\newblock \url{http://www.ctan.org/pkg/nomencl}.

\bibitem{Talbot16:Glossaries}
Nicola L.~C. Talbot.
\newblock {\em User Manual for glossaries.sty v4.44}, December 2019.
\newblock \url{http://www.ctan.org/pkg/glossaries}.

\bibitem{Carlisle04:Textcase}
David Carlisle.
\newblock {\em The \textsl{textcase} package}, October 2004.
\newblock \url{http://www.ctan.org/pkg/textcase}.

\end{thebibliography}



\begin{thebibliography}{21}


\ifx \showCODEN    \undefined \def \showCODEN     #1{\unskip}     \fi
\ifx \showDOI      \undefined \def \showDOI       #1{#1}\fi
\ifx \showISBNx    \undefined \def \showISBNx     #1{\unskip}     \fi
\ifx \showISBNxiii \undefined \def \showISBNxiii  #1{\unskip}     \fi
\ifx \showISSN     \undefined \def \showISSN      #1{\unskip}     \fi
\ifx \showLCCN     \undefined \def \showLCCN      #1{\unskip}     \fi
\ifx \shownote     \undefined \def \shownote      #1{#1}          \fi
\ifx \showarticletitle \undefined \def \showarticletitle #1{#1}   \fi
\ifx \showURL      \undefined \def \showURL       {\relax}        \fi
\providecommand\bibfield[2]{#2}
\providecommand\bibinfo[2]{#2}
\providecommand\natexlab[1]{#1}
\providecommand\showeprint[2][]{arXiv:#2}

\bibitem[\protect\citeauthoryear{Adair, Li, Yang, and Yu}{Adair
  et~al\mbox{.}}{2017}]%
        {adair2017progress}
\bibfield{author}{\bibinfo{person}{Bill Adair}, \bibinfo{person}{Chengkai Li},
  \bibinfo{person}{Jun Yang}, {and} \bibinfo{person}{Cong Yu}.}
  \bibinfo{year}{2017}\natexlab{}.
\newblock \showarticletitle{Progress Toward “the Holy Grail”: The Continued
  Quest to Automate Fact-Checking}. In \bibinfo{booktitle}{\emph{Computation+
  Journalism Symposium, Evanston}}.
\newblock


\bibitem[\protect\citeauthoryear{Bawden and Robinson}{Bawden and
  Robinson}{2009}]%
        {bawden2009dark}
\bibfield{author}{\bibinfo{person}{David Bawden} {and} \bibinfo{person}{Lyn
  Robinson}.} \bibinfo{year}{2009}\natexlab{}.
\newblock \showarticletitle{The dark side of information: overload, anxiety and
  other paradoxes and pathologies}.
\newblock \bibinfo{journal}{\emph{Journal of information science}}
  \bibinfo{volume}{35}, \bibinfo{number}{2} (\bibinfo{year}{2009}),
  \bibinfo{pages}{180--191}.
\newblock


\bibitem[\protect\citeauthoryear{Cohen, Hamilton, and Turner}{Cohen
  et~al\mbox{.}}{2011}]%
        {cohen2011computational}
\bibfield{author}{\bibinfo{person}{Sarah Cohen}, \bibinfo{person}{James~T
  Hamilton}, {and} \bibinfo{person}{Fred Turner}.}
  \bibinfo{year}{2011}\natexlab{}.
\newblock \showarticletitle{Computational journalism}.
\newblock \bibinfo{journal}{\emph{Commun. ACM}} \bibinfo{volume}{54},
  \bibinfo{number}{10} (\bibinfo{year}{2011}), \bibinfo{pages}{66--71}.
\newblock


\bibitem[\protect\citeauthoryear{Devlin, Chang, Lee, and Toutanova}{Devlin
  et~al\mbox{.}}{2018}]%
        {devlin2018bert}
\bibfield{author}{\bibinfo{person}{Jacob Devlin}, \bibinfo{person}{Ming-Wei
  Chang}, \bibinfo{person}{Kenton Lee}, {and} \bibinfo{person}{Kristina
  Toutanova}.} \bibinfo{year}{2018}\natexlab{}.
\newblock \showarticletitle{Bert: Pre-training of deep bidirectional
  transformers for language understanding}.
\newblock \bibinfo{journal}{\emph{arXiv preprint arXiv:1810.04805}}
  (\bibinfo{year}{2018}).
\newblock


\bibitem[\protect\citeauthoryear{Diakopoulos, Naaman, and
  Kivran-Swaine}{Diakopoulos et~al\mbox{.}}{2010}]%
        {diakopoulos2010diamonds}
\bibfield{author}{\bibinfo{person}{Nicholas Diakopoulos}, \bibinfo{person}{Mor
  Naaman}, {and} \bibinfo{person}{Funda Kivran-Swaine}.}
  \bibinfo{year}{2010}\natexlab{}.
\newblock \showarticletitle{Diamonds in the rough: Social media visual
  analytics for journalistic inquiry}. In \bibinfo{booktitle}{\emph{2010 IEEE
  Symposium on Visual Analytics Science and Technology}}. IEEE,
  \bibinfo{pages}{115--122}.
\newblock


\bibitem[\protect\citeauthoryear{Finnas}{Finnas}{2018}]%
        {newsworthy_slideshow}
\bibfield{author}{\bibinfo{person}{Jens Finnas}.}
  \bibinfo{year}{2018}\natexlab{}.
\newblock \showarticletitle{The hard parts about automating journalism}.
\newblock \bibinfo{journal}{\emph{Google Slides}} (\bibinfo{year}{2018}).
\newblock
\urldef\tempurl%
\url{https://docs.google.com/presentation/d/1iXHDg-h0sWjaLFC3ku6rJKD6FVFA-7naIy8i_LDcof8}
\showURL{%
\tempurl}


\bibitem[\protect\citeauthoryear{Galtung and Ruge}{Galtung and Ruge}{1965}]%
        {galtung1965structure}
\bibfield{author}{\bibinfo{person}{Johan Galtung} {and}
  \bibinfo{person}{Mari~Holmboe Ruge}.} \bibinfo{year}{1965}\natexlab{}.
\newblock \showarticletitle{The structure of foreign news: The presentation of
  the Congo, Cuba and Cyprus crises in four Norwegian newspapers}.
\newblock \bibinfo{journal}{\emph{Journal of peace research}}
  \bibinfo{volume}{2}, \bibinfo{number}{1} (\bibinfo{year}{1965}),
  \bibinfo{pages}{64--90}.
\newblock


\bibitem[\protect\citeauthoryear{Goswami}{Goswami}{2015}]%
        {goswami2015analysing}
\bibfield{author}{\bibinfo{person}{Shubham Goswami}.}
  \bibinfo{year}{2015}\natexlab{}.
\newblock \showarticletitle{Analysing effects of information overload on
  decision quality in an online environment}.
\newblock \bibinfo{journal}{\emph{Journal of Management Research}}
  \bibinfo{volume}{15}, \bibinfo{number}{4} (\bibinfo{year}{2015}),
  \bibinfo{pages}{231--245}.
\newblock


\bibitem[\protect\citeauthoryear{Hagel~III and Singer}{Hagel~III and
  Singer}{1999}]%
        {hagel1999private}
\bibfield{author}{\bibinfo{person}{John Hagel~III} {and} \bibinfo{person}{Marc
  Singer}.} \bibinfo{year}{1999}\natexlab{}.
\newblock \showarticletitle{Private lives}.
\newblock \bibinfo{journal}{\emph{The McKinsey Quarterly}} \bibinfo{number}{1}
  (\bibinfo{year}{1999}), \bibinfo{pages}{6}.
\newblock


\bibitem[\protect\citeauthoryear{Han, Ning, and Peng}{Han
  et~al\mbox{.}}{2019}]%
        {han2019joint}
\bibfield{author}{\bibinfo{person}{Rujun Han}, \bibinfo{person}{Qiang Ning},
  {and} \bibinfo{person}{Nanyun Peng}.} \bibinfo{year}{2019}\natexlab{}.
\newblock \showarticletitle{Joint Event and Temporal Relation Extraction with
  Shared Representations and Structured Prediction}.
\newblock \bibinfo{journal}{\emph{arXiv preprint arXiv:1909.05360}}
  (\bibinfo{year}{2019}).
\newblock


\bibitem[\protect\citeauthoryear{Ho and Tang}{Ho and Tang}{2001}]%
        {ho2001towards}
\bibfield{author}{\bibinfo{person}{Jinwon Ho} {and} \bibinfo{person}{Rong
  Tang}.} \bibinfo{year}{2001}\natexlab{}.
\newblock \showarticletitle{Towards an optimal resolution to information
  overload: an infomediary approach}. In \bibinfo{booktitle}{\emph{Proceedings
  of the 2001 international ACM SIGGROUP conference on supporting group work}}.
  \bibinfo{pages}{91--96}.
\newblock


\bibitem[\protect\citeauthoryear{Joulin, Grave, Bojanowski, and Mikolov}{Joulin
  et~al\mbox{.}}{2016}]%
        {joulin2016bag}
\bibfield{author}{\bibinfo{person}{Armand Joulin}, \bibinfo{person}{Edouard
  Grave}, \bibinfo{person}{Piotr Bojanowski}, {and} \bibinfo{person}{Tomas
  Mikolov}.} \bibinfo{year}{2016}\natexlab{}.
\newblock \showarticletitle{Bag of tricks for efficient text classification}.
\newblock \bibinfo{journal}{\emph{arXiv preprint arXiv:1607.01759}}
  (\bibinfo{year}{2016}).
\newblock


\bibitem[\protect\citeauthoryear{Koltay}{Koltay}{2017}]%
        {koltay2017bright}
\bibfield{author}{\bibinfo{person}{Tibor Koltay}.}
  \bibinfo{year}{2017}\natexlab{}.
\newblock \showarticletitle{The bright side of information: ways of mitigating
  information overload}.
\newblock \bibinfo{journal}{\emph{Journal of Documentation}}
  (\bibinfo{year}{2017}).
\newblock


\bibitem[\protect\citeauthoryear{Lee, So, Lee, Leung, and Chan}{Lee
  et~al\mbox{.}}{2018}]%
        {lee2018social}
\bibfield{author}{\bibinfo{person}{Paul~SN Lee}, \bibinfo{person}{Clement~YK
  So}, \bibinfo{person}{Francis Lee}, \bibinfo{person}{Louis Leung}, {and}
  \bibinfo{person}{Michael Chan}.} \bibinfo{year}{2018}\natexlab{}.
\newblock \showarticletitle{Social media and political partisanship--A
  subaltern public sphere’s role in democracy}.
\newblock \bibinfo{journal}{\emph{Telematics and Informatics}}
  \bibinfo{volume}{35}, \bibinfo{number}{7} (\bibinfo{year}{2018}),
  \bibinfo{pages}{1949--1957}.
\newblock


\bibitem[\protect\citeauthoryear{Liu, Ott, Goyal, Du, Joshi, Chen, Levy, Lewis,
  Zettlemoyer, and Stoyanov}{Liu et~al\mbox{.}}{2019}]%
        {liu2019roberta}
\bibfield{author}{\bibinfo{person}{Yinhan Liu}, \bibinfo{person}{Myle Ott},
  \bibinfo{person}{Naman Goyal}, \bibinfo{person}{Jingfei Du},
  \bibinfo{person}{Mandar Joshi}, \bibinfo{person}{Danqi Chen},
  \bibinfo{person}{Omer Levy}, \bibinfo{person}{Mike Lewis},
  \bibinfo{person}{Luke Zettlemoyer}, {and} \bibinfo{person}{Veselin
  Stoyanov}.} \bibinfo{year}{2019}\natexlab{}.
\newblock \showarticletitle{Roberta: A robustly optimized bert pretraining
  approach}.
\newblock \bibinfo{journal}{\emph{arXiv preprint arXiv:1907.11692}}
  (\bibinfo{year}{2019}).
\newblock


\bibitem[\protect\citeauthoryear{Mayer, Heiser, and Lonn}{Mayer
  et~al\mbox{.}}{2001}]%
        {mayer2001cognitive}
\bibfield{author}{\bibinfo{person}{Richard~E Mayer}, \bibinfo{person}{Julie
  Heiser}, {and} \bibinfo{person}{Steve Lonn}.}
  \bibinfo{year}{2001}\natexlab{}.
\newblock \showarticletitle{Cognitive constraints on multimedia learning: When
  presenting more material results in less understanding.}
\newblock \bibinfo{journal}{\emph{Journal of educational psychology}}
  \bibinfo{volume}{93}, \bibinfo{number}{1} (\bibinfo{year}{2001}),
  \bibinfo{pages}{187}.
\newblock


\bibitem[\protect\citeauthoryear{McIntyre}{McIntyre}{2016}]%
        {mcintyre2016makes}
\bibfield{author}{\bibinfo{person}{Karen McIntyre}.}
  \bibinfo{year}{2016}\natexlab{}.
\newblock \showarticletitle{What makes “good” news newsworthy?}
\newblock \bibinfo{journal}{\emph{Communication Research Reports}}
  \bibinfo{volume}{33}, \bibinfo{number}{3} (\bibinfo{year}{2016}),
  \bibinfo{pages}{223--230}.
\newblock


\bibitem[\protect\citeauthoryear{Miller}{Miller}{2018}]%
        {miller2018news}
\bibfield{author}{\bibinfo{person}{Judith Miller}.}
  \bibinfo{year}{2018}\natexlab{}.
\newblock \showarticletitle{News deserts: No news is bad news}.
\newblock \bibinfo{journal}{\emph{Urban policy 2018}} (\bibinfo{year}{2018}),
  \bibinfo{pages}{59--76}.
\newblock


\bibitem[\protect\citeauthoryear{{\"O}stgaard}{{\"O}stgaard}{1965}]%
        {ostgaard1965factors}
\bibfield{author}{\bibinfo{person}{Einar {\"O}stgaard}.}
  \bibinfo{year}{1965}\natexlab{}.
\newblock \showarticletitle{Factors influencing the flow of news}.
\newblock \bibinfo{journal}{\emph{Journal of peace Research}}
  \bibinfo{volume}{2}, \bibinfo{number}{1} (\bibinfo{year}{1965}),
  \bibinfo{pages}{39--63}.
\newblock


\bibitem[\protect\citeauthoryear{Pedregosa, Varoquaux, Gramfort, Michel,
  Thirion, Grisel, Blondel, Prettenhofer, Weiss, Dubourg, Vanderplas, Passos,
  Cournapeau, Brucher, Perrot, and Duchesnay}{Pedregosa et~al\mbox{.}}{2011}]%
        {scikit-learn}
\bibfield{author}{\bibinfo{person}{F. Pedregosa}, \bibinfo{person}{G.
  Varoquaux}, \bibinfo{person}{A. Gramfort}, \bibinfo{person}{V. Michel},
  \bibinfo{person}{B. Thirion}, \bibinfo{person}{O. Grisel},
  \bibinfo{person}{M. Blondel}, \bibinfo{person}{P. Prettenhofer},
  \bibinfo{person}{R. Weiss}, \bibinfo{person}{V. Dubourg}, \bibinfo{person}{J.
  Vanderplas}, \bibinfo{person}{A. Passos}, \bibinfo{person}{D. Cournapeau},
  \bibinfo{person}{M. Brucher}, \bibinfo{person}{M. Perrot}, {and}
  \bibinfo{person}{E. Duchesnay}.} \bibinfo{year}{2011}\natexlab{}.
\newblock \showarticletitle{Scikit-learn: Machine Learning in {P}ython}.
\newblock \bibinfo{journal}{\emph{Journal of Machine Learning Research}}
  \bibinfo{volume}{12} (\bibinfo{year}{2011}), \bibinfo{pages}{2825--2830}.
\newblock


\bibitem[\protect\citeauthoryear{White and Dorman}{White and Dorman}{2000}]%
        {white2000confronting}
\bibfield{author}{\bibinfo{person}{Marsha White} {and} \bibinfo{person}{Steve~M
  Dorman}.} \bibinfo{year}{2000}\natexlab{}.
\newblock \showarticletitle{Confronting information overload}.
\newblock \bibinfo{journal}{\emph{Journal of School Health}}
  \bibinfo{volume}{70}, \bibinfo{number}{4} (\bibinfo{year}{2000}),
  \bibinfo{pages}{160--160}.
\newblock


\end{thebibliography}
\bibliographystyle{ACM-Reference-Format}

\end{document}